\documentclass[letterpaper]{article} 
\usepackage{aaai23}  
\usepackage{times}  
\usepackage{helvet}  
\usepackage{courier}  
\usepackage[hyphens]{url}  
\usepackage{graphicx} 
\usepackage{natbib}  
\usepackage{caption} 
\usepackage{algorithm}
\usepackage{algorithmic}

\usepackage{subcaption}
\usepackage{booktabs}
\usepackage{multirow}
\usepackage{graphicx}
\usepackage{etoolbox}
\usepackage{amsmath}
\usepackage{xcolor}
\usepackage{color}
\newcommand{\sun}[1]{{\color{black} #1}}
\newcommand{\shi}[1]{{\color{blue} #1}}
\newrobustcmd{\B}{\bfseries}

\newtheorem{proposition}{Proposition}

\usepackage{newfloat}
\usepackage{listings}
\DeclareCaptionStyle{ruled}{labelfont=normalfont,labelsep=colon,strut=off} 
\floatstyle{ruled}
\newfloat{listing}{tb}{lst}{}
\floatname{listing}{Listing}
\title{Scalable Optimal Multiway-Split Decision Trees with Constraints}
\author {
    Shivaram Subramanian\equalcontrib,
    Wei Sun\equalcontrib
}
\affiliations {
    IBM Research\\1101 Kitchawan Road\\Yorktown Heights, New York 10598, USA\\
    \{subshiva, sunw\}@us.ibm.com
}

\usepackage{bibentry}

\begin{document}

\maketitle

\begin{abstract}
There has been a surge of interest in learning optimal decision trees using mixed-integer programs (MIP) in recent years, as heuristic-based methods do not guarantee optimality and find it challenging to incorporate constraints that are critical for many practical applications.    However, existing MIP methods that build on an \emph{arc-based} formulation do not scale well as the number of binary variables is in the order of $\mathcal{O}(2^dN)$, where $d$ and $N$ refer to the depth of the tree and the size of the dataset. Moreover, they can only handle sample-level constraints and linear metrics. In this paper, we propose a novel \emph{path-based} MIP formulation where the number of decision variables is independent of 
$N$. We present a scalable column generation framework to solve the MIP optimally. 
Our framework produces a multiway-split tree which is more interpretable than the typical binary-split trees due to its shorter rules. Our method can handle nonlinear metrics such as F1 score and incorporate a broader class of constraints. We demonstrate its efficacy with extensive experiments. We present results on datasets containing up to 1,008,372  samples while existing MIP-based decision tree models do not scale well on data beyond a few thousand points. We report superior or competitive results compared to the state-of-art MIP-based methods with up to a 24X reduction in runtime.
\end{abstract}

\section{Introduction}

Decision trees are among the most popular machine learning models as the tree structure is visually easy to understand. 
As learning an optimal  decision tree is NP-hard \cite{laurent1976constructing}, popular algorithms such as CART \cite{breiman1984classification}, ID3 
\cite{quinlan1986induction} and C4.5 \cite{quinlan2014c4} rely on greedy heuristics to construct trees. Motivated by the heuristic nature of the traditional methods, there have been many efforts across different fields to learn optimal decision trees (ODT), e.g., dynamic programming  \cite{lin2020generalized}, constraint programming \cite{verhaeghe2020learning}, Boolean satisfiability \cite{narodytska2018learning}, itemset mining \cite{aglin2020learning}. 
In particular, recent advances in modern optimization has facilitated a nascent stream of research that leverages mixed-integer programming  (MIP) to train globally optimal
trees with constraints  \cite{bertsimas2017optimal,aghaei2019learning,verwer2019learning, aghaei2020learning} - this is the methodology we are focusing on in this paper.

\begin{figure}[h]
  \centering
\includegraphics[width=.9\linewidth]{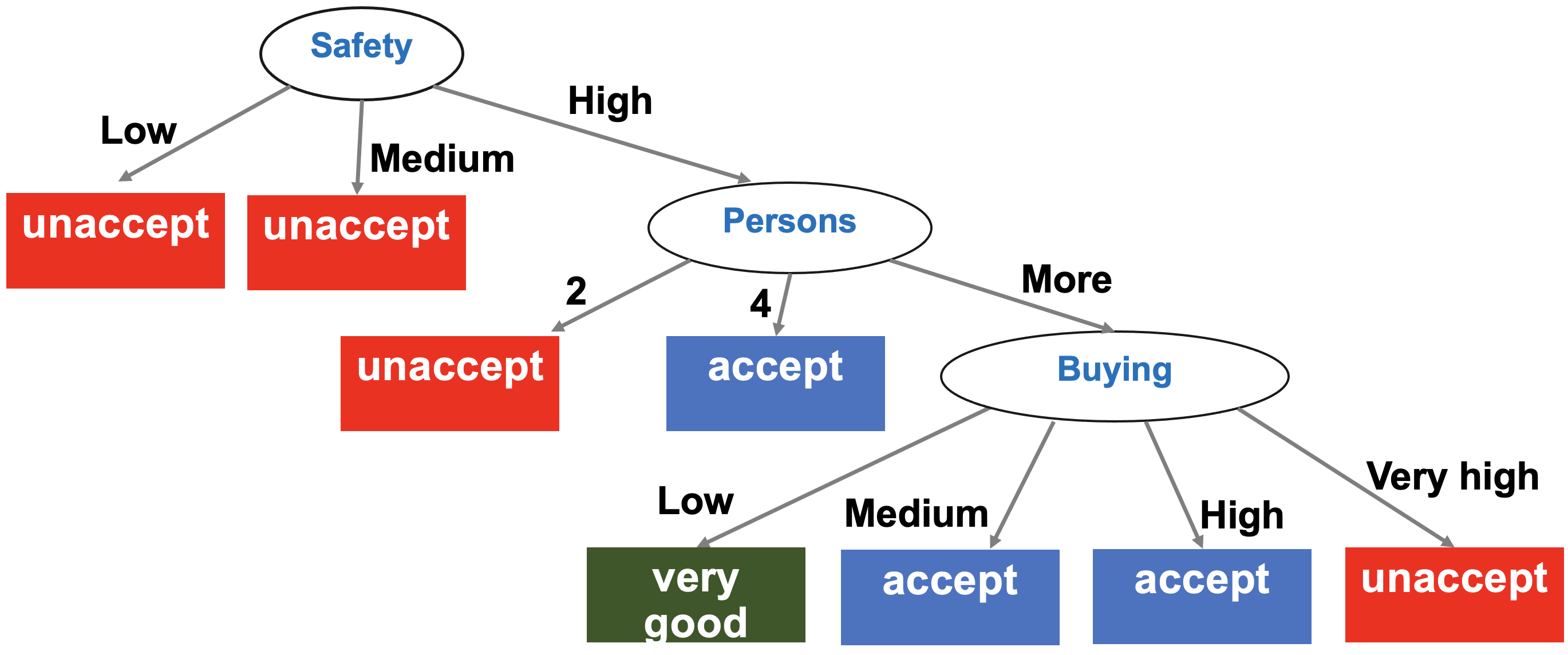}
\caption{An example of OMT trained on \textsf{car-evaluation}.}\label{fig_OMT}
\end{figure}

Prior MIP-based methods rely on an \emph{arc-based} formulation that require a large number of binary decision variables to identify splitting conditions at branch nodes as well as label and sample  assignments to leaf nodes. 
This approach has several drawbacks: (1) the optimization problem becomes easily intractable as the number of binary variables and constraints increases linearly with training data.  Hence, experiments are typically restricted to datasets with no more than a few thousand samples. 
(2) Prior MIP frameworks can only handle linear metrics.  
In many applications, it is  desirable to consider nonlinear metrics, e.g., F1-score is preferred over accuracy to evaluate  machine learning models trained on imbalanced datasets. (3) With the arc-based formulation, it is challenging to impose constraints on individual decision rules and feature combinations. One such example occurs in the medical field, where doctors have to arrive at an appropriate diagnosis while taking into account the costs of medical tests  \cite{lomax2013survey}. 
(4) The vast majority of the decision tree literature focuses on binary-split trees, where each node can have at most two child nodes. 
Multiway-split trees (see Figure~\ref{fig_OMT} for an example) whose branching condition may contain several values are often more intuitive and comprehensible \cite{fulton1995efficient}. 


In this paper, we attempt to address  these prior limitations by proposing a scalable MIP-based framework to train constrained \emph{optimal multiway-split trees} (OMT).
Our contributions are five-fold.

\textbf{Novel path-based MIP formulation} Unlike arc-based MIP, we explicitly model decision rules or paths in a tree. 
The high level idea is as follows - we first define a feature graph that admits every possible combination
of input features as feasible decision rule candidates. The objective of the MIP is to identify a constrained subset of rules from this potentially huge rule space to minimize the prediction error. The feature graph naturally embeds the hierarchical structure of a tree, and we add a constraint to model another property of decision trees, i.e., each sample can only be assigned to a single rule. 

\textbf{Scalable algorithm}
Unlike existing MIP ODTs whose binary decision variables  are typically in the order of $\mathcal{O}(2^dN)$, where $d$ and $N$ refer to the depth of the tree and the size of training data,  the 
number of binary decision variables in our formulation is independent of 
$N$. It equals the number of candidate decision rules defined in the feature graph. While this number 
can also be excessive in the worst case, we employ column generation (CG) to dynamically generate a relatively small number of candidates to recover a near-optimal solution.
While
existing MIP approaches can manage no more than a few thousand training samples, an initial implementation of our proposed CG  algorithm effectively solved a $10^6$ sample dataset. 

\textbf{Flexible framework} 
Our  framework can be applied to both classification and regression  settings. The path-based model allows us to incorporate
any path-level metric, including nonlinear ones, which enter the MIP as input parameters.
We also show how a broader class of constraints including path-level, and attribute-level constraints,  which are difficult to express within existing arc-based formulations, can be easily incorporated into our CG framework.

\textbf{Enhanced interpretability with multiway-splits} To the best of our knowledge, we are the first to train optimal multiway-split trees for prediction tasks, where a node may have more than two child nodes. 
Since a feature seldom appears more than once in any root-to-leaf path, multiway-split trees are easier to comprehend than their binary counterparts \cite{fulton1995efficient}. 

\textbf{Extensive numerical results} 
We conduct extensive computational studies on public datasets and report the first MIP-based results for large datasets containing up to 1,008,372 samples, which are an order of magnitude larger than the prior datasets analyzed in the MIP-based ODT literature. We show that our framework is either competitive or improves upon the state-of-the-art MIP methods in terms of out-of-sample accuracy and achieves up to a 24-fold reduction in runtime.



\section{Related literature}
The discrete nature of decisions involved in training a decision tree and recent algorithmic advances in integer optimization have inspired a burgeoning body of literature that utilize a MIP formulation to construct optimal decision trees \cite{bertsimas2017optimal,verwer2019learning,aghaei2019learning,NEURIPS2020_1373b284,aghaei2020learning,gunluk2021optimal}. 
Most of the existing MIP methods build on top of the Optimal Classification Tree (OCT) first introduced in \citealp{bertsimas2017optimal}, where decisions about the split condition at each node, label assignment for each leaf node, and the routing of each data point from the root node to a leaf node are made. 
Different flavors of optimal decision trees have been proposed in the literature, e.g., discrimination-aware trees \cite{aghaei2019learning}, trees with combinatorial splitting rules \cite{gunluk2021optimal}, multivariate trees \cite{NEURIPS2020_1373b284}.  

As the tractability of the original formulation is limited by  data size, subsequent efforts  sought to improve the efficiency of the approach.
The BinOCT approach of \citealp{verwer2019learning} employs a binary encoding method to model the threshold selection at branch nodes to 
reduce the number of binary decision variables needed. However, both the OCT and BinOCT  use “big-M” constraints which may weaken the underlying LP relaxations, leading to poor performance of the branch-and-bound method. Most notably, a recent work  proposes FlowOCT \cite{aghaei2020learning}, a strong max-flow based MIP formulation with binary data. 
Its formulation yields a tighter underlying LP relaxation and outperforms prior methods in many instances. 
Nevertheless,  existing methods' tractability is limited by data size and tree depth as the arc-based formulation uses binary variables to assign samples to nodes. This motivates us to seek a completely different approach, 
by explicitly modeling the paths used to construct a tree. 

While the worst-case number of rules can be huge with high-dimensional data, we show that our problem can be solved efficiently via column generation (CG). CG has been successful in solving large-scale discrete optimization models in many domains including vehicle routing (\citealp{chen2006dynamic}), crew scheduling (\citealp{subramanian2008effective, bront2009column}),  and supply chain management, among others (\citealp{xu2019solving}).
Utilizing large-scale optimization techniques for MIP-based ODTs has been attempted previously. 
\citealp{aghaei2020learning} show that  FlowOCT  can be solved by Benders’ decomposition (row generation). Its subproblem involves solving a max-flow problem for \emph{every} sample potentially, limiting
results to datasets having a few thousand samples.

\section{Problem formulation}

We consider a dataset which consists of $N$  samples, $\{(x_i,y_i)\}_{i=1}^N$, where $x_i \in \mathcal{X}^k$ are features which are assumed to be categorical. We will discuss  how to handle  numerical features in Section~\ref{sect_cumulative_binning}, and present experiments on datasets with both categorical and numerical features in Section~\ref{sect_experiments}. 
$y_i$ is the outcome, which can be a discrete label for classification or a continuous quantity for regression.

A binary-split tree of depth $d$ can have at most $2^d$ leaf nodes. In a multiway-split  tree, each node may have more than two children. Thus, we use the depth of a tree $d$, as well as the number of leaf nodes $l$, which are user-specified parameters, to describe such a tree. An example of a multiway-split tree with $d=3$ and $l=8$ is shown in Figure \ref{fig_OMT}.



In our framework, we explicitly model individual decision paths from the root to the leaf nodes. We begin by defining a feature graph 
that contains every possible combination of input features, followed by introducing a MIP optimization problem to identify a subset of paths to form the tree. 


\subsection{Feature graph}\label{sect_rule_space}


We consider an acyclic multi-level digraph, $G(V, E)$, where each  feature indicates a level in the graph, represented by multiple nodes corresponding to its distinct feature values. 
Nodes of a feature are fully connected to nodes in the next level. The graph includes a source and sink node. A decision rule is defined as a path 
from the source to the sink node. 

\begin{figure}[h]
  \centering
\includegraphics[width=0.8\linewidth]{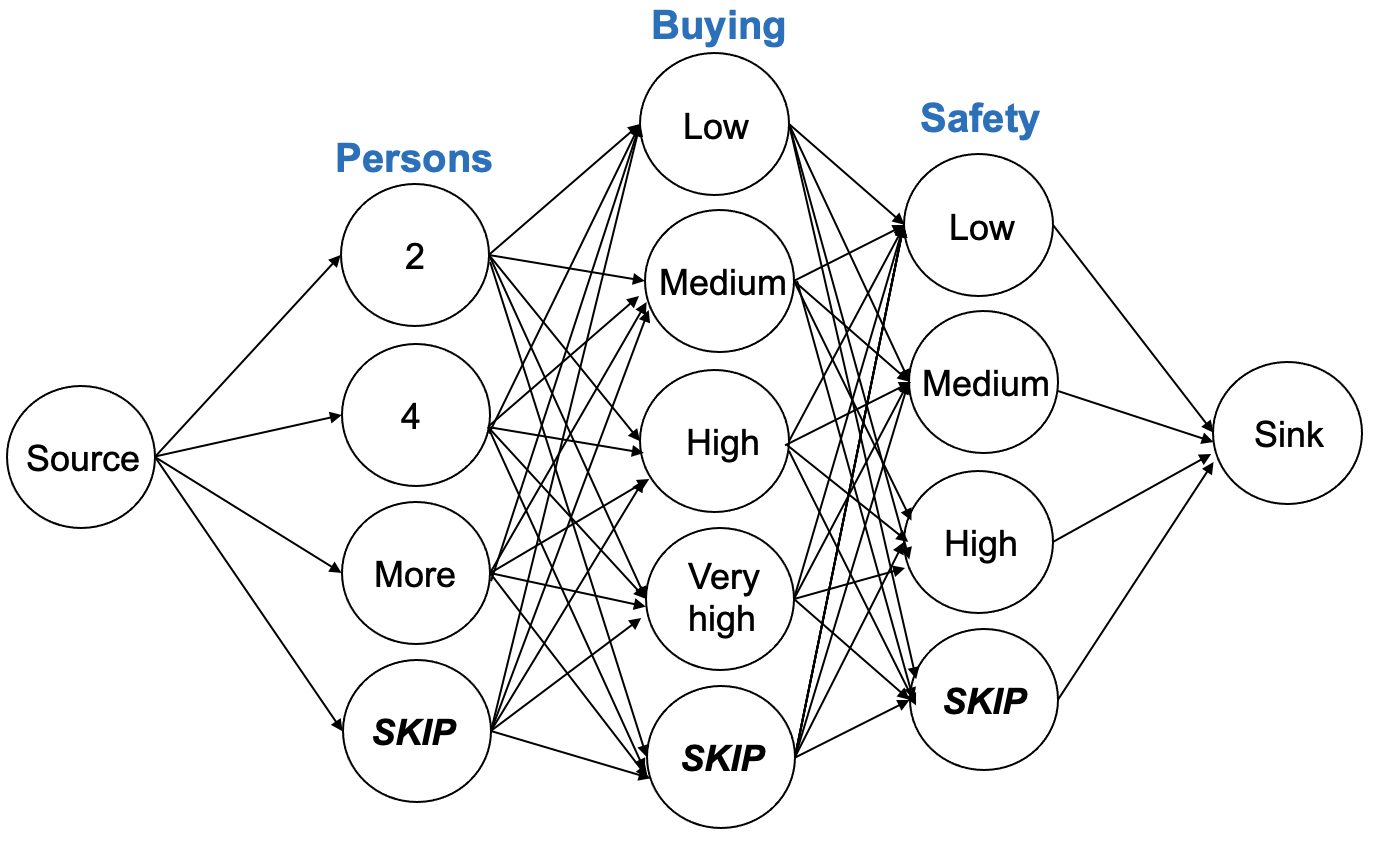}
\caption{The feature graph for \textsf{car-evaluation}
}\label{feature_space}
\end{figure}

For each feature, we introduce a dummy node \emph{SKIP}. 
If  a path goes through \emph{SKIP} node of a feature, it means that the particular feature is not used in a decision rule. 
%
As  \emph{SKIP} nodes allow paths to ignore  features, paths on this acyclic graph  represent  all possible feature combinations, defining the full search space $\mathcal{P}$ that one may need to consider to construct an optimal tree.  Figure~\ref{feature_space} illustrates an example of a feature graph using three features from \textsf{car-evaluation}, a UCI dataset \cite{dua2017uci}, which records  assessment of cars based on criteria such as ``Persons'' (number of seats), ``Buying'' (purchase cost) and ``Safety''. 

Denote $\eta$ as the maximum unique feature values associated with a feature. The following result implies that enumerating through the space $\mathcal{P}$ may become impractical for high-dimensional datasets. 
\begin{proposition}\label{prop_feature_space}
    The search space $\mathcal{P}$ is  $\mathcal{O}(\eta^k).$
\end{proposition}
Later in Section \ref{sect_algorithm}, we describe our algorithm which uses a large-scale optimization technique to intelligently search through the space, and only paths that can improve the current solution are generated on the fly. 
\subsection{Path-Based MIP formulation}\label{sect_formulation}
We denote a decision rule as $j \in\{ 1,\cdots,L\}$, where $L=|\mathcal{P}|$.  Let $S_j \in [N]$ be the subset of observations which fall into rule $j$. 
Denote $\xi_j$ as a user-specified metric associated with rule $j$, e.g., misclassification error in rule $j$ in a classification task, squared loss or absolute loss for regression.

Let $z_j$ be a binary decision variable which indicates whether rule $j$ is selected (1) or not (0), for $j = 1,\cdots,L$. 
{We define non-negative slack variables $s_i$  and a positive penalty cost $c_i$ for each sample $i$. }
A MIP formulation to determine an optimal multiway-split tree ({OMT}) can be written as follows, 
\begin{align}
\min\quad & \sum_{j=1}^L \xi_jz_j +\sum_{i=1}^N c_is_i\nonumber\\
\textrm{s.t.} \quad & \sum_{j=1}^L a_{ij}z_j + s_i = 1, \quad\forall i=1,\cdots,N\label{constraint_coverage}\\
&\sum_{j=1}^L z_j \leq l \label{capacity}\\
  &z_j\in\{1,0\}, \quad \forall j=1,\cdots,L \nonumber\\ &s_i \geq 0, \quad \forall i=1,\cdots,N \nonumber
\end{align}
The input parameter $a_{ij}=1$ if sample $i$ satisfies the conditions specified in rule $j$, and 0 otherwise.  
%
While several rules may contain the same data sample, {the set partitioning constraint (\ref{constraint_coverage}) 
with a sufficiently large penalty $c_i$ ensures that each sample is ultimately assigned to exactly one rule.}
The cardinality constraint (\ref{capacity}) stipulates that no more than $l$ rules are active in the optimal solution $\mathbf{z}^*$. 


It is well-known that set partitioning problems are \textit{NP}-hard \cite{hartmanis1982computers}. 
Thus, while optimal solutions can be obtained in practice for moderate sized  instances \cite{atamturk1996combined}, the  problem may easily become intractable to solve directly as the cardinality of feasible rules grows exponentially with the feature space. We will present an efficient algorithm to overcome this computational challenge in the next section.



\section{Column generation}\label{sect_algorithm} 

Column generation (CG) is a classical technique to solve large MIPs for which it is not practical to explicitly generate  all columns (variables) of the problem \cite{lubbecke2005selected}. 
Specifically, we first consider a \emph{restricted master program} (\textsf{RMP}) version of \textsf{OMT}, where we  1) consider only a subset of paths $\hat{L}$, which is typically much smaller than $L$ and is determined dynamically, and 2) relax the integrality constraints on $z_j$ to $0\leq z_j\leq 1$ for all $j=1,\cdots,\hat{L}$.

Denote the dual variables   associated with the set partitioning constraints in (\ref{constraint_coverage}) and the cardinality constraint in (\ref{capacity})  as ${\lambda}_i$ and $\mu$ respectively. The dual formulation of RMP can be written as follows, 
\begin{align}
\quad\max\quad & \sum_{i=1}^N \lambda_i + n\mu\nonumber\\
\textrm{s.t.} \quad & \sum_{i=1}^N a_{ij}\lambda_i+\mu\leq \xi_j, \quad \forall j=1,\cdots,\hat{L}\label{dual_constraint}\\
& \mu\leq0, \quad \lambda_i\leq c_i, \quad\forall i = 1,\cdots, N 
 \nonumber
\end{align}
Note that since $z_j\leq 1$ is implied by Eq (\ref{constraint_coverage}), we are only left with $z_j \geq 0$ in the primal \textsf{RMP}.


Based on the dual feasibility constraint for path $j$ in  (\ref{dual_constraint}),  the reduced cost for path $j$ is equal to
\begin{equation}\small
    rc_j =  \xi_j -\left(\sum_{i=1}^N a_{ij}\lambda_i + \mu\right). \label{eq_reduced cost}
\end{equation}

When RMP is solved to optimality, dual feasibility is guaranteed only for the rules included in $\hat{L}$. 
A path violating the dual constraint  (\ref{dual_constraint}) has a negative reduced cost and must be added to the RMP for the next iteration. 
To  identify paths that maximally violate Eq (\ref{dual_constraint}), we need to solve $\min rc_j$. 
We define the \emph{subproblem}, as the search for some $K$ paths having the most negative reduced cost,  
which is a 
\emph{K-shortest path problem} (KSP) over the feature graph $G$,  
where the cost on a path is its reduced cost defined in (\ref{eq_reduced cost}). If no violating path exists, we obtain dual feasibility, and the CG has converged. 


The high-level view of the CG procedure is as follows: we start
by solving RMP with an empty set of candidate rules with $\hat{L}=\emptyset$, which sets slack variables $s_i$ to 1.0 and yields an initial dual solution. 
Next, we use these dual values to solve the subproblem to identify up to $K$  candidate rules having the most negative reduced cost. They are added to $\hat{L}$ and the RMP is re-optimized to obtain a new primal-dual solution, and the process iterates. 
The CG procedure converges to an optimal solution of the LP relaxation of \textsf{OMT} when dual feasibility is achieved, i.e., $rc_j \geq 0$  for all $j = 1,\cdots,M$. 
After the CG procedure has converged, the binary restrictions are reimposed on $\mathbf{z}$ and we solve the resultant Master-MIP.


The CG procedure can be implemented exactly for provable optimality, or approximately to obtain a near-optimal decision tree. Prior works \cite{barnhart1998branch} have proposed a branch-and-price technique to solve the Master-MIP to provable optimality, which also requires the subproblem to be solved exactly 
\cite{desrosiers2005primer}.

For simplicity and replicability, we employ a CG heuristic that solves the resultant Master-MIP directly using a standard optimization package \cite{cplex2020}.
{As noted in \citealp{lubbecke2005selected}, the role of the subproblem is to provide a potential column or to prove that none exists. 
Since any negative reduced cost path contributes to this aim,
we employ a heuristic subproblem algorithm that is a path-dependent adaptation of a $K$-shortest path method  for acyclic graphs with additive arc costs \cite{horne1980finding}. \sun{Our KSP method is similar to the resource-constrained shortest path heuristics used in CG applications \cite{desrosiers2005primer,desaulniers2006column}} and has polynomial-time complexity $\mathcal{O}(K\eta|V|N log N)$. Details are given in the appendix. In our experiments, the CG procedure terminates when the dual solution converges to within a tolerance or we reach a maximum iteration limit. }



{To preserve interpretability, shallow trees are often preferred, where the depth $d$ is small. The following result shows how this constraint dramatically shrinks the CG search space, especially for high-dimensional datasets.
\begin{proposition}
  The constrained search space is   $\mathcal{O}\big({k \choose d}\eta^d\big)$.
\end{proposition}
Comparing this result to the unconstrained setting shown in Proposition~\ref{prop_feature_space}, we see that restricting each path to contain no more than $d$ features significantly reduces the total number of feasible paths. In other words, the worst-case value for $\hat{L}\ll L$. Considering an example with {$d=\eta=2$,  $\hat{L}$ is bounded by $\mathcal{O}(k^2)$ versus $L$ is $\mathcal{O}(2^k)$.
{Prior MIP models are restricted to a fixed-depth binary tree representation which requires categorical attributes to be encoded into a generic set of binary features. Doing so prevents them from exploiting the beneficial hierarchical feature graph structure used in our approach.} 
}
\section{Flexible framework}

\subsection{Nonlinear metrics}\label{sect_nonlinear_metrics}
Nonlinear metrics such as the F1-score, Matthews correlation coefficient, and Fowlkes–Mallows index, are often used to evaluate the performance of machine learning models trained on imbalanced data \cite{nonlinearmetric2021}. 
Existing MIP-based decision trees only consider linear metrics.   In our approach, once the feature graph is defined, we know the samples $S_j$ which satisfy this rule and we can compute the metric associated with this rule in the subproblem. 
These nonlinear metrics may be represented as $\xi_j$ and enter RMP as an input parameter in the objective as shown in Section \ref{sect_formulation}. They can also enter RMP as constraints and we provide an example of incorporating F1-score in the appendix. This transformative modeling capability of CG is invaluable when solving  optimization problems that are highly nonlinear and nonconvex in their original form. 

\subsection{Constraints enforcement}\label{sect_constraints}
Existing optimal classification trees in the literature have been extended to incorporate constraints to address fairness and imbalance issues.
In this section, we show how our method provides an elegant and unified framework to handle constraints, including those which existing arc-based MIP formulations cannot efficiently manage.

\paragraph{Path-level constraints}
Prior MIP-based decision trees typically manage \emph{sample-level} constraints, e.g., constraining precision or recall conditioned on samples' class labels \cite{aghaei2020learning,gunluk2021optimal}, or fairness metrics such as statistical parity conditioned on sensitive features \cite{aghaei2019learning,aghaei2020learning}. 
However, none of the existing MIP-based methods are able to efficiently handle constraints at \emph{path-level}, as the notion of ``path'' is not explicitly defined in the arc-based formulation. 

Consider the example of cost-sensitive decision trees, which is motivated by the medical domain. Often, doctors must arrive at a diagnosis by taking into account the economic constraints faced by a patient when different test options involve a tradeoff between accuracy and measurement cost \cite{lomax2013survey, nunez1991use}. 
Denoting the cost associated with each decision rule $j$ which consists of several medical tests as $\rho_j$  and the budget as $C$, we specify the following constraint: $\rho_jz_j\leq C$ for all $j=1,\cdots, L$. This constraint which ensures that the final selected diagnostic methods (with $z_j=1)$ are staying within the budget,  can be processed within the KSP subproblem. 
Additional examples 
can be found in the appendix. 


In general, \emph{path-level} constraints can be expressed as a set of polyhedral inequalities on $\mathbf{z}$, i.e., $\sum_{j=1}^L\rho_{mj}z_j\geq q_m, \forall m=1,\cdots, M.$
Let $\tau_m \geq 0$ denote the dual variables corresponding to these constraints. Then, the reduced cost $rc_j$ for path $j =  \xi_j -\left(\sum_{i=1}^N a_{ij}\lambda_i + \mu +\sum_{m=1}^M \rho_{mj}\tau_m \right).$
These constraints influence the subproblem through the resultant reduced costs to ensure that paths that are more likely to be feasible are added to the RMP. 
 \begin{table*}[t] \small
  \centering
\begin{tabular}{lllrccccc} 
\toprule
       dataset & $N$  &$d$ &         OMT &         OCT &      BinOCT &     FlowOCT &        CART \\
\midrule
 soybean-small &    47& 4 & 0.883±0.139 & 0.944±0.048 &  0.833±0.22 & 0.944±0.083 &    \B 1.0±0.0 \\
 soybean-small &    47&  5 &  0.95±0.075 & 0.972±0.096 &  0.722±0.21 & 0.972±0.083 &   \B  1.0±0.0 \\
       monks-3 &    122&  4 & 0.987±0.008 &  0.99±0.015 & 0.988±0.011 &  0.99±0.011 & \B 0.993±0.007 \\
       monks-3 &    122&   5 & 0.987±0.008 & 0.978±0.014 & 0.983±0.015 &  0.99±0.012 & \B 0.993±0.007 \\
       monks-1 &    124& 4 &    \B 1.0±0.0 &   \B 1.0±0.029 &     \B 1.0±0.0 &    \B 1.0±0.03 & 0.806±0.064 \\
       monks-1 &    124&  5 &     \B 1.0±0.0 & 0.935±0.142 &     \B 1.0±0.0 &   \B  1.0±0.03 & 0.787±0.042 \\
    hayes-roth &    132&   4 &  0.75±0.075 &  0.75±0.038 & 0.642±0.138 &   \B0.8±0.066 &  0.55±0.087 \\
    hayes-roth &132&   5 &  0.77±0.089 &  0.75±0.076 & 0.575±0.066 & \B 0.817±0.029 & 0.708±0.058 \\
       monks-2 &     169&  4 & 0.603±0.045 & \B 0.662±0.031 & 0.581±0.027 & \B 0.662±0.023 & 0.598±0.033 \\
       monks-2 &     169&  5 & \B 0.779±0.03 &  0.662±0.05 &  0.607±0.03 & 0.662±0.055 & 0.651±0.055 \\   
house-votes-84 &    232&  4 & 0.962±0.046 & \B  0.971±0.01 &  0.914±0.03 &  \B 0.971±0.01 &   0.96±0.02 \\
house-votes-84 &    232&  5 & 0.955±0.026 &  \B 0.971±0.01 &   0.96±0.01 &  \B 0.971±0.01 &   0.96±0.02 \\
         spect &  267&    4 & \B 0.834±0.04 & 0.801±0.086 & 0.746±0.065 & 0.791±0.079 & 0.731±0.026 \\
         spect &   267&  5 & \B 0.803±0.039 &  0.791±0.06 & 0.721±0.048 & 0.796±0.085 &  0.731±0.03 \\
 breast-cancer &    277&  4 &  0.74±0.046 & 0.724±0.038 & 0.662±0.091 & \B 0.743±0.079 &  0.69±0.058 \\
 breast-cancer &    277&  5 & 0.669±0.094 &   \B 0.738±0.0 & 0.567±0.157 & 0.676±0.036 & 0.671±0.049 \\
 balance-scale &    625&  4 & \B 0.781±0.016 & 0.747±0.048 & 0.707±0.006 &  0.699±0.01 & 0.769±0.033 \\
 balance-scale &    625&  5 & \B 0.772±0.012 &  0.735±0.01 &   0.565±0.1 &  0.72±0.045 & 0.762±0.013 \\
   tic-tac-toe &   958&   4 & \B 0.802±0.039 & 0.776±0.073 & 0.786±0.021 & 0.757±0.032 & 0.758±0.019 \\
   tic-tac-toe &   958&   5 &  \B 0.82±0.061 & 0.711±0.025 & 0.812±0.029 &  0.788±0.05 & 0.778±0.046 \\
car-evaluation &    1728&  4 & \B 0.879±0.009 & 0.796±0.076 &  0.848±0.01 & 0.823±0.016 & 0.842±0.029 \\
car-evaluation &    1728& 5 &  \B 0.91±0.005 & 0.742±0.041 & 0.815±0.052 &   0.8±0.016 & 0.857±0.019 \\
      kr-vs-kp &    3196& 4 &  \B 0.96±0.006 & 0.847±0.094 & 0.938±0.012 &  0.94±0.011 &  0.94±0.011 \\
      kr-vs-kp &    3196&  5 & \B 0.968±0.003 & 0.652±0.098 & 0.847±0.164 & 0.946±0.057 &  0.94±0.011 \\
\bottomrule
\end{tabular}
\caption{Mean ± standard deviation of out of sample accuracy on the small/medium datasets. 
}\label{table_small_data}
\end{table*}
\paragraph{Attribute-level constraints}
Existing MIP methods are also challenged by \emph{attribute-level} constraints that represent complex nonlinear conditions involving several features that can not be abstracted efficiently into linear constraints, e.g., disallowing certain feature combinations. 
These constraints are easily handled within the KSP subproblem as a feasibility check while extending a partial path to the next node in $G$.
A practical example of an attribute-level requirement is the need to preserve attribute hierarchy. 
{For example, in} the medical domain, doctors may perform temperature checks or blood tests before proceeding to more advanced tests. Enforcing such hierarchy makes the decision tree more reliable and comprehensible for domain experts \cite{nanfack2022constraint}.  This can be easily achieved
by appropriately arranging the nodes in the feature graph.

\subsection{Cumulative binning on numerical features} \label{sect_cumulative_binning}
For tree-based approaches, numerical input are typically handled via thresholding. For example, consider a numerical feature with values in [0, 1] being divided into 3 intervals, e.g., [0, 0.33), [0.33, 0.67) and [0.67, 1.0]. One can transform this numerical feature to a categorical feature with 3 values, and create 3 nodes representing this feature in the graph. This approach may be limiting as a binary-split tree can branch on conditions such as $x\leq 0.67$ or $x> 0.3$. To address this limitation, we employ \emph{cumulative binning}, where intervals can be overlapping. We create additional nodes representing intervals {[0, 0.67),  [0.33, 1.0]}, yielding a total of 5 nodes. 
 The coverage constraints (\ref{constraint_coverage}) 
 ensure that the final rule set does not contain overlapping samples. In our experiments, quantile discretization is used to create intervals. More generally, with $\kappa$ intervals, cumulative binning results in $\mathcal{O}(\kappa^2$) nodes, i.e, a gain in a rule's expressiveness at the cost of higher computational effort.

\section{Experiments}\label{sect_experiments}
While {OMT} is a general framework which can
also be used in regression settings,
we focus on  classification tasks in our experiments as most of the existing ODTs are classification trees.  
We group our experiments by dataset size, where 
  \textit{small/medium} datasets are loosely defined as having a few thousand samples, 
  and a dataset is considered \textit{large} if it has many thousands of samples and several hundred binary features  \cite{dash2018boolean}. 
 
To benchmark  our approach \textsf{OMT}, we implement the following MIP-based methods, i.e., \textsf{FlowOCT} \cite{aghaei2020learning}, \textsf{BinOCT} \cite{verwer2019learning} and \textsf{OCT} \cite{bertsimas2017optimal}. 
 Although \textsf{CART} cannot produce constrained decision rules, we still include it as a baseline.
 As the benchmarks produce binary-split trees of a given depth $d$, we construct a comparable multiway-split tree via the cardinality constraint that  restricts the number of leaf nodes $l$ to be at most $2^d$, i.e., $l=2^d$, and limit the rule length to $d$. 


CPLEX 20.1 \cite{cplex2020} was used to solve the MIP-based methods.
\textsf{CART} was trained using scikit-learn \cite{pedregosa2011scikit} using default hyper-parameters. 
The minimum number of samples per rule for \textsf{OMT} was set to 1\% of the training data. The maximum value for $K$ was set to 1000 in the KSP for all instances except the large dataset experiments, where it was reduced to 100 to stay within the RAM limit. We set a maximum CG iteration limit of 40 and a $\hat{L}$ limit of 10,000. All experiments were run on an Intel 8-core i7 PC with 32GB RAM. Details on the experiment setup can be found in the appendix.

\subsection{Small/medium datasets}
We evaluate the same 12 classification datasets from the UCI repository \cite{dua2017uci} that have been used in \textsf{FlowOCT} \cite{aghaei2020learning}, which is considered the state-of-the-art ODT. 
We closely follow the experiment setup in \citealp{aghaei2020learning} to construct decision trees with  $d\in\{2,3,4,5\}$. 
We create 5 random splits for each dataset into training (50\%), validation (25\%), and test sets (25\%). 
A time limit of 20 minutes is imposed on each experiment, in contrast to the one hour limit used  in \citealp{aghaei2020learning}.  

Table \ref{table_small_data} reports the 
achieved out-of-sample accuracy averaged over five splits for $d\in\{4,5\}$ (the complete results across all depths can be found in the appendix). Best  accuracy in a given row is reported in \textbf{bold}. Among 48 (data, depth) combinations, \textsf{OMT} dominates other methods in 56.3\% of them, \textsf{FlowOCT} 33.3\%, \textsf{OCT} 29.2\%, \textsf{BinOCT} 8.3\%, and \textsf{CART} 14.6\% (including ties). These results demonstrate that our \textsf{OMT} method achieves competitive and often superior results compared to MIP-based binary-split ODT models. 

\begin{figure}[h]
  \centering
 \includegraphics[width=1\linewidth]{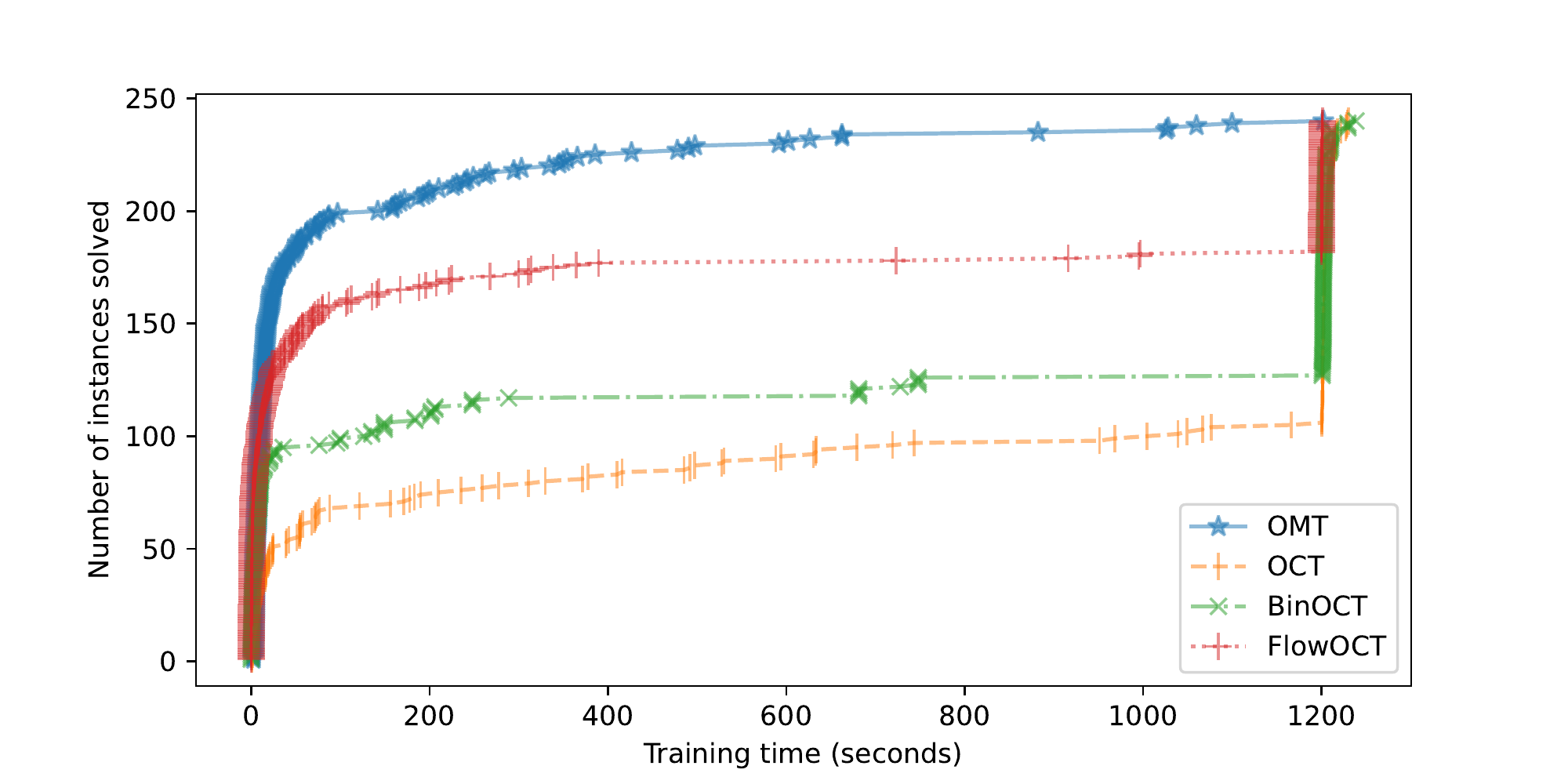}
  \caption{Runtime performance on a total of 240 instances}\label{fig_runtime1}
\end{figure}
Next, we analyze the solution time taken by different MIP-based methods.  We define an ``instance''  as a unique (dataset, depth, data split) combination. With 5 random data splits, there are $12\times4\times5=240$ instances in total.
In Figure \ref{fig_runtime1}, the x-axis shows the solution time  in seconds, while the y-axis shows the number of instances solved. With the MIP-based ODT benchmarks, when the solution time is under the time limit, implying the instance is solved to near optimality.    
Figure \ref{fig_runtime1} shows that \textsf{OMT} solves most instances before the 20 minutes time limit, followed by  \textsf{FlowOCT},  \textsf{BinOCT} and  \textsf{OCT}. Specifically,  188 out of 240 instances was solved under 1 minute by \textsf{OMT}, in contrast to 137, 95 and 59 instances by \textsf{FlowOCT},  \textsf{BinOCT} and  \textsf{OCT} respectively;  230 instances out of 240  was solved under 10 minutes by \textsf{OMT}, 
compared to 175, 118 and 90 instances by the benchmarks. 
The bulk of the unsolved instances by the time limit is the medium datasets with $d\geq3$. 

Table \ref{table_runtime} reports the solution times of different MIP methods with respect to tree depth.  The solution time of unsolved instances is capped at 20 minutes. Due to the skewed distributions contributed by huge discrepancies in terms of data size and feature size, we report the median values. As CART finishes all instances in seconds, we omit the results in the table. {In general, CART's solutions can be infeasible as it ignores many of the constraints we consider in this work.}

Table \ref{table_runtime} shows that the state-of-the-art MIP-based benchmark, \textsf{FlowOCT} 
takes the least time per instance among all methods including \textsf{OMT} at $d=2$, However, it experiences a nearly 20X increase in runtime as $d$ increases from 2 to 3, which grows by another 4.5X when $d$ = 4. We believe this is an under-statement as the solution time of many unsolved instances at higher depth are capped at 20 minutes. 
A further increase in runtime is observed in \textsf{OCT} and \textsf{BinOCT} whose binary variables increase exponentially with $d$. In contrast, for \textsf{OMT}, the runtime increases are merely 1.4X, 0.68X, 0.4X as $d$ increases from 2 to 5. 
Overall, our speedup over \textsf{FlowOCT} at $d=4$ and 5 is 10.3X and 24.6X respectively. The exact OMT speedup is likely to be even  higher if we compare runtime to achieve the same solution quality. 

{The strong CG performance can also be explained by examining the gap between the objective value achieved by the Master-MIP ($\nu_{IP}$) and the final RMP ($\nu_{LP}$), i.e.,  $\Delta = (\nu_{IP} - \nu_{LP})/\nu_{IP}$. A table that reports on this MIP-LP gap with respect to $d$ is included in the appendix. 
In particular, the median gap is no more than 0.3\% across all depths, suggesting that the path-based LP relaxation is a relatively strong approximation of the nonconvex discrete OMT model. This tight gap also underscores the advantage of converging to a relatively small subset of high quality paths from which an effective decision tree can be distilled using a standard MIP solver. } 


\begin{table}\small
 \centering
\begin{tabular}{rrrrr}
\toprule
 $d$ &    OMT &  FlowOCT &  BinOCT &     OCT \\
\midrule
     2 &    3.9  &    1.9  &     6.1  &   33.9 \\
     3 &   9.3  &  32.5&  727.8  & 1200.4  \\
     4 & 15.6  & 177.0  & 1201.9  & 1202.5 \\
     5 & 21.9 & 561.4  & 1203.5  & 1205.9  \\
\bottomrule
\end{tabular} 
\caption{Median runtime (seconds) for MIP-based methods}
\label{table_runtime}\end{table}

\begin{table*}[t]
  \small
\makebox[\textwidth][c]{
 \begin{tabular}{cccccccccc}
\toprule
   & &   &   & \multicolumn{5}{c}{Univariate} & \multicolumn{1}{c}{Multivariate} \\
\cmidrule(lr){5-9}\cmidrule(lr){10-10}
  dataset &      $N$ &  $k$ &  $d$ &   OMT &   OCT &  BinOCT &  FlowOCT &  CART &   S1O \\
\midrule
pendigits &   7494 & 16 &  2 & \B0.395$^*$ & 0.348 &   0.291 &    0.329 & 0.352/0.362 & 0.389 \\
pendigits &   7494 & 16 &  3 & \B0.685$^*$ & 0.517 &   0.347 &    0.459 & 0.558/0.579 & 0.625 \\
    avila &  10430 & 10 &  2 & 0.475 & 0.461 &   0.096 &    \B0.509 & 0.501/0.503 & 0.526$^*$ \\
    avila &  10430 & 10 &  3 & 0.524 & 0.417 &   0.403 &    0.504 & \textbf{0.527}/0.535 & 0.558$^*$ \\
      EEG &  14980 & 14 &  2 & \B0.658 & 0.602 &   0.584 &    0.648 & 0.624/0.586 & 0.665$^*$ \\
      EEG &  14980 & 14 &  3 & \B0.690$^*$ & 0.572 &   0.493 &    0.649 & 0.659/0.642 & 0.665 \\
     HTRU &  17898 &  8 &  2 & 0.973 & 0.977 &   0.705 &    0.956 & \textbf{0.978}/0.973 & 0.978$^*$ \\
     HTRU &  17898 &  8 &  3 & 0.977 & 0.978 &   0.552 &    0.973 & \textbf{0.979}/0.981$^*$ & 0.979 \\
  shuttle &  43500 &  9 &  2 & \B0.968$^*$ & 0.821 &   0.285 &    0.920 & 0.939/0.938 & 0.940 \\
  shuttle &  43500 &  9 &  3 & 0.984 & 0.793 &   0.390 &    0.914 & \textbf{0.996}/0.997 & 0.995$^*$ \\
     skin & 245057 &  3 &  2 & 0.875$^*$ & 0.899 &   0.774 &    0.802 & \textbf{0.907}/0.806 & 0.863 \\
     skin & 245057 &  3 &  3 & \B0.967$^*$ & 0.793 &   0.855 &    0.801 & 0.965/0.871 & 0.949 \\
\bottomrule
\end{tabular}
}
\caption{Out of sample accuracy, using the large datasets from \cite{NEURIPS2020_1373b284}. }
\label{table_zhu_dataset_full} 
\end{table*}%

\subsection{Large datasets}
We test our method on the six largest datasets analyzed in the MIP-based ODT literature \cite{NEURIPS2020_1373b284}. 
We follow the same experiment setup described earlier but limit runtime to one hour to account for the larger data size. 
As in \citealp{NEURIPS2020_1373b284}, we focus on $d=\{2,3\}$. Table \ref{table_zhu_dataset_full} summarizes the average out-of-sample accuracy of different MIP approaches. The \emph{left} entries under column ``CART" shows the \textsf{CART} results that we obtain. 
None of the arc-based MIP methods are able to obtain an optimal solution on any of the 12 large instances at $d=3$ within the time limit, with \textsf{FlowOCT} winning in one instance, and no wins for \textsf{OCT} and \textsf{BinOCT}. On the other hand, \textsf{OMT} dominates 6 out of 12, and
 \textsf{CART} wins 5 out of 12 cases. {While the quality improvement over \textsf{CART} is not dramatic, the latter is unable to handle constraints, which is a key practical feature of our approach.}  {Meanwhile, we improve upon the average misclassification error achieved by \textsf{FlowOCT} by 8.4\%.} 

Table \ref{table_zhu_dataset_full} also includes a column ``S1O", taken  from \citealp{NEURIPS2020_1373b284}. For ease of comparison, we  include their \textsf{CART} values shown as the right entries under column ``CART" in Table \ref{table_zhu_dataset_full}. We want to point out that it is not quite fair to compare the two: Firstly,  \textsf{OMT} is a univariate tree with axis-parallel splits, whereas \textsf{S1O} is a multivariate (oblique) tree, wherein splits at a node use multiple variables, or hyperplanes. These multivariate splits tend to be much stronger than univariate splits as shown in \cite{bertsimas2017optimal}, at the expense of losing interpretability.  
Furthermore, the time limit for \textsf{S1O} reported in \cite{NEURIPS2020_1373b284} was four hours, while we limit \textsf{OMT} to an hour ({the average \textsf{OMT} runtime per instance is 413 and 835 seconds at $d = 2$ and 3, respectively}).  Finally, instead of running on the raw data as in \textsf{OMT},  \textsf{S1O} employs a LP-based data-selection preprocessing step. 
Nevertheless, we benchmark \textsf{OMT} against \textsf{S1O} (and their CART results)  in Table \ref{table_zhu_dataset_full}, and the best achieved accuracy is marked with a star ($*$). Of the 12  cases, \textsf{OMT} still wins 6, while \textsf{S1O} and \textsf{CART} wins 5 and 2 cases respectively. 
These results highlight our method's ability to produce competitive results in a relatively short training time, and without sacrificing interpretability. 

Lastly, we stress-test our method by analyzing challenging datasets in the UCI repository that are an order of magnitude larger than those reported for prior optimal ODT methods. We increase the degree of difficulty in two dimensions: More samples (up to one million) and more raw features (up to $k=175$). 
For such large datasets, we found a negligible change in solution quality for different random seeds, so we solved each instance once using a fixed random seed and a six-hour time limit and report the results in Table \ref{table_additional_data}.  As none of the prior optimal ODTs could process such large data, we compare the achieved solution quality to CART. 
Note via proposition~\ref{prop_feature_space} that setting $d = 2$ for the \textit{crop-mapping} dataset ($\eta=4$}) yields $\hat{L} \leq 16\times175^2$ versus $L=4^{175}$. We improve upon CART's test accuracy in 6 of the 8 instances with an average runtime of 3.7 hours per large instance. 


\begin{table}[t]
\centering\small
\begin{tabular}{lrrrrr} 
\toprule
     dataset &      $N$ &   $k$ &  $d$ &   OMT &  CART \\
\midrule
       MiniBooNE & 130065 &  50 &  2 & \B0.850 & 0.835 \\
       MiniBooNE & 130065 &  50 &  3 & \B0.873 & 0.865 \\
crop mapping & 325834 & 175 &  2 & \B0.769 & 0.679 \\
crop mapping & 325834 & 175 &  3 & \B0.841 & 0.797 \\
   covertype & 581012 &  54 &  2 & 0.667 & \B0.668 \\
   covertype & 581012 &  54 &  3 & \B0.684 & 0.677 \\
   susy & 1008372
 &  18 &  2 & 0.744 & \B0.748 \\
   susy & 1008372
 &  18 &  3 & \B0.760 & 0.755 \\
\bottomrule
\end{tabular}
\caption{Out-of-sample accuracy on additional large datasets}\label{table_additional_data} 
\end{table}

\section{Conclusion}
In this paper, we  propose a scalable and flexible mixed-integer optimization framework based on column generation (CG) to identify 
optimal multiway-split decision trees with constraints. 
We present a novel path-based MIP formulation for decision trees where the number of columns or binary variables is independent of the training data size. 
Our method can generate  both classification and regression trees by minimizing any of the commonly used nonlinear error metrics while still solving a linear MIP model. As each categorical feature only enters a rule once, multiway-split trees are easier to comprehend than their binary-split counterpart. 
A first ``plain vanilla'' CG implementation is tested on several public datasets ranging up to a million data samples. We are able to achieve a performance that is comparable to or superior than those achieved by the state-of-art optimal classification tree methods in the literature while consuming only a small fraction of their runtime. 
Furthermore, our CG based formulation is able to seamlessly handle a wide variety of practical constraints that cannot be efficiently managed by prior ODT models or popular heuristics like \textsf{CART}.

The current framework has to discretize numerical features. One way to handle numerical features \emph{without} discretization is to employ a preprocessing step like BinOCT [Verwer \& Zhang 2019], which we leave as future work. 
 


\bibliography{ref}

\end{document}


\maketitle

\section{Additional Details for Section 3: Problem Formulation}

\subsection{Proof for Proposition 1}
\begin{proof}
Given a $k$-dimensional input data $x^k$, denote $\eta_f$ as the number of unique feature values associated with the $f^{th}$ feature.  The number of paths from the source to the sink is given by $|\mathcal{P}|=O\left(\prod_{f=1}^k \eta_f\right)=O\left(\eta^k\right)$, for some constant $\eta$. 
\end{proof}

\subsection{Feature graph}

While the solution quality (in terms of the objective value of the MIP) is independent of feature order and different feature orders lead to different decision trees (via different feature graphs), the interpretability of the solution may depend on the order. In Section 5, we give an example in the medical field where a certain feature order representing the sequence of diagnostic tests is preferred by doctors. Our approach provides an easy way to enforce such a structure in the rules by simply arranging features according to the preferred order in the feature graph.  In another scenario, one may first run a black-box prediction model to learn the feature importance (e.g., SHAP scores \cite{lundberg2017unified}), and arrange the features in descending order of importance. This is particularly relevant for high-dimensional datasets where $k$ is large. 
Using sorted features, one can find a good solution quickly by exploring the more important feature combinations first.

\subsection{Optimal decision tree via solving OMT}
Decision trees are different from rule sets -- the latter consists of unordered rules which may overlap, whereas a decision tree has a hierarchical structure of features and each example is covered by exactly a single rule. The feature graph provides an embedded hierarchical structure of features.  Thus, when combined with the set coverage constraint which enforces each sample can only be assigned to one rule, the optimal solution to the mixed integer optimization problem, OMT, identifies  an optimal multiway-split decision tree. 

Once we obtain the tree,  the label of a leaf node is determined in the same way as in traditional decision trees, i.e., with a classification tree, the label of a leaf node/path is the majority class of samples in that node.




\section{Additional Details for Section 4: Column Generation }

Column generation (CG) is nowadays a prominent method to cope with a huge number of variables. We refer readers to \cite{lubbecke2005selected} for an excellent survey on column generation, where numerous integer programming CG applications are described.

\subsection{Proof for Proposition 2}
\begin{proof}
When  we impose a constraint on tree depth $d$,  we restrict to consider the paths that contain $d$ features, which are in the order of $\eta^d$ by Proposition 1. As we are also choosing $d$ out of  $k$ features, giving rise to ${k \choose d}$ combinations. Thus, the feasible path set for a tree of depth $d$ is in $O\left( {k \choose d} \eta^d   \right)$.
\end{proof}

Compare this result  to Proposition 1 where we do not limit rule length, we see that having a constraint on the tree depth vastly reduces the total number of feasible paths. Additional constraints such as the minimum number of samples per path further reduce the number of feasible paths (e.g., in our experiments, we require at least 1\% of training data for the selected rules). We discuss in details in Section 5 how such constraints can be easily incorporated in our approach during the KSP subproblem. 

\subsection{Subproblem Heuristic}
While an exact subproblem solution is required to achieve provable optimality \cite{barnhart1998branch}, as pointed out in \cite{lubbecke2005selected}, there is often no need to solve the subproblem exactly  -- the role of the subproblem is to provide a column with a negative reduced cost or to prove that none exists. It is important to see that any column with negative reduced cost contributes to this aim. Our proposed KSP method is a polynomial-time heuristic path-dependent adaptation of the $K$-shortest path method for acyclic networks having arc costs \cite{horne1980finding, eppstein1998finding}.

On a high level, the KSP algorithm starts from the source node and proceeds in a fixed order according to the feature graph, visiting every node and extending up to the $K$ best reduced-cost paths at each node to the nodes of the next feature layer. {This algorithm differs from the standard shortest path algorithms as the path costs are not additive over arcs. }When a path is extended to include the next feature node, metrics $\xi_j$ such as misclassification error  and the resulting reduced cost are re-computed.


More specifically, given a feature graph $G$, we label the nodes from 1 through $|V|$, starting from the source and proceeding to the sink, where node $1$ represents the source node and node $|V|$ is the sink node. With the exception of the sink node, for each node $t$, denote $C(t)$ as its children nodes in the next feature layer. We denote $\Pi_t$ as the set of (up to) K shortest paths (i.e., with the lowest reduced cost) at node $t$. 

For each path $j$ in the feature graph, recall that $S_j$ represents the samples that fall into rule $j$. 
Once $S_j$ is identified, we can compute the loss metric $\xi_j$ 
as well as the reduced cost $rc_j$ according to Equation (4) in the paper. Denote $\phi_j$ as an ordered index set of the visited feature nodes in path $j$. 

For the algorithm, we define the following two operators: 

$\Leftarrow$: \textit{Insert} a path $\phi_j$ into the path set $\Pi_t$ and retain the $K$ best paths in $\Pi_t$. 

$\oplus$: \textit{Extend} a path $\phi_j$ by adding node $t$ to it and create a new path $\phi_j^{new}$. 

For example, in the feature graph shown in Figure 2, we may extend the path $\phi_j:=$\{Source, Persons-2, Buying-High\} to node \{Safety-Medium\} to create a new path $\phi_{j^{new}}:=$\{Source, Persons-2, Buying-High, Safety-Medium\}. 
This operation consists of three sub-tasks:

\begin{enumerate}

\item	Filter $S_j$ to obtain a subset of samples, $S_{j^{new}}$, which is covered by the new path $\phi_{j^{new}}$. 
    
\item	Verify the feasibility of $\phi_{j^{new}}$ to attribute and path constraints and return $\O$ if infeasible.
    
\item Compute the loss metric $\xi_{j^{new}}$ and the resulting reduced cost $rc_{j^{new}}$ = $\xi_{j^{new}} - (\sum_{i\in S_{j^{new}}} \lambda_i +\mu) $.

\end{enumerate}

The pseudo-code for the KSP subproblem is shown in Algorithm~\ref{alg:cap}.  The nodes are examined in sequence from the source to the sink. In Line 2 to 5, for every node $t$, we examine its KSP partial path list, $\Pi_t$. If a path has a negative reduced cost, it is immediately inserted into $\Pi_{|V|}$, i.e., the KSP path list at the sink node. In Line 6 to 8, we extend a path to the nodes in the next feature layer. The output of the algorithm is the KSP list at its sink node, $\Pi_{|V|}$.

\begin{algorithm}
\caption{KSP subproblem}\label{alg:cap}
\noindent\textbf{Input}: $K,\lambda, \mu$  

\noindent\textbf{Output}: $\Pi_{|V|}$ \-\hspace{3cm}  

\noindent\textbf{Initialize}: $\Pi_1 \gets \{1\}, \Pi_t \gets \O$, for  $t=2,\cdots, |V|$
\begin{algorithmic}[1]
\For{$t$ = 1 to $|V|$}  
    \ForAll{$j$ in  $\Pi_t$} \-\hspace{1.9cm} 
        \If{ $rc_j < 0$}
            \State $\Pi_{|V|} \Leftarrow \phi_j$ 
        \EndIf
    \ForAll{$m \in C(t)$}
            \State $\Pi_m \Leftarrow \phi_j \oplus m$ 
    \EndFor
    \EndFor
\EndFor
\end{algorithmic}
\end{algorithm}

Our extensive experiments show that this heuristic works well in terms of runtime and achieving near-optimal solution quality. 
Resource-constrained shortest path heuristics (\cite{desrosiers2005primer, desaulniers2006column} similar to our KSP procedure have been applied in large-scale industrial applications. For example \cite{subramanian2008effective} apply a KSP-like subproblem algorithm within a large-scale airline crew scheduling optimizer on graphs having several thousand nodes. Although the network sizes they analyze are considerably larger than the biggest feature graph we report on, the \textit{Crop Mapping} Dataset (175 categorical features) we analyze (see the experiments section) has by far the largest feature set in the MIP-based ODT literature to the best of our knowledge.

\subsection{Complexity analysis}

{For exposition,  consider solving OMT for a classification task. }
We begin by analyzing the worst case runtime of the KSP subproblem. First, note that we visit every node and perform no more than $K\eta$ path extensions from each node. Next, the runtime complexity of every such path extension $\phi_j \oplus m$ is obtained by examining the computation time for each of the three sub-tasks:  

Sub-task i) Filtering: computing the intersection of the sets of data samples belonging to the path, and the next node, can be performed in O$(N log N)$ time using an efficient sorting algorithm.

Sub-task ii) Assuming we only have a maximum rule length $d$ and a minimum sample size constraint and we store the length and number of data samples belonging to the path, feasibility can be verified in $\mathcal{O}(1)$ time.

Sub-task iii) Computing the loss metric $\xi_j$ requires us to count the number of data samples of the path that belong to each class. The misclassification cost $\xi_j$ is simply the difference between the number of data samples of the path and the sample count for the most popular class. The dual values ($\lambda,\mu$) of the data samples have to be summed to recalculate the reduced cost. The computation time is linear in the data sample size: $\mathcal{O}(|S_j|)$ time, or $\mathcal{O}(N)$ in the worst case.

The runtime is dominated by the first subtask, giving us a worst case $\mathcal{O}(N log N)$ time per path extension, which yields an overall polynomial runtime bound of $\mathcal{O}(K\eta|V|N log N)$ for KSP.  

{
RMP, being a linear program, is also solvable in polynomial time \cite{bazaraa2008linear}, and therefore every CG iteration of RMP and KSP runs in polynomial time. }

{Note that the Master-MIP that is solved after the CG procedure is an NP-Hard problem due the presence of the binary variables that inject non-convexity. The number of constraints in the RMP or the MIP increase in proportion to the number of data samples but are not a limiting factor as they are convex and processed efficiently within the (polynomial time solvable) LP subproblems solved during the MIP solver's branch-and-bound procedure \cite{wolsey1999integer}. This ability to efficiently handle constraints are also why MIP-based methods are a valuable solution approach to ODTs}.

\subsection{Limitations and potential enhancements}
 There are several limitations regarding the current implementation. For simplicity and replicability, we chose to employ standard off-the-shelf methods (e.g. CPLEX Barrier and MIP solver for RMP and Master-MIP).  For instance, the final Master-MIP problem can be solved to optimality with more advanced methods such as branch-and-price instead of a standard solver. The RMP implementation can be enhanced using subgradient methods, which are usually much faster\cite{subramanian2008effective}. Also, rather than start from scratch with no columns, the CG procedure can be ``warm-started'' using the rules obtained by running CART.
 
 The existing KSP procedure  is also a ``plain vanilla'' implementation for easy reproducibility, and can be enhanced in several ways. For example, it can be implemented in parallel easily by dividing the path search among different processors. Next, the nodes in the feature graph can be arranged based on the SHAP-score for the corresponding features, which can be obtained from any blackbox prediction model. This can accelerate KSP and CG convergence to good quality solutions much earlier.

\section{Additional Details for Section 5: Flexible Framework}
\subsection{Incorporate nonlinear metrics in constraints}
To model a F1-score at the sample-level or dataset-level, we first compute path-level metrics such as true positives (tp), false positive (fp), false negatives (fn). To be more precise, suppose we want to impose a constraint that F1 at the dataset-level is above a certain threshold $\delta$, that is, 

$$F1=\frac{\sum_j tp_jz_j}{\sum_j (tp_j + 0.5(fp_j + fn_j))z_j}\geq \delta.$$
We can then rewrite this as a linear constraint and add it to the RMP, i.e., $$\sum_j tp_jz_j\geq \delta\sum_j \left(tp_j + 0.5(fp_j + fn_j)\right)z_j.$$

Such a constraint will influence the subproblem via a modified reduced cost to ensure that paths that are more likely to be feasible are added to the RMP. The same idea can be applied to Matthews correlation coefficient and Fowlkes–Mallows index. With the proposed framework, we can also impose path-level constraints which we discuss in the next section. 

\subsection{Fairness constraints}

Prior fairness metrics are defined as  \emph{sample-level} constraints, i.e., statistical parity conditioned on certain sensitive features such as gender or race \cite{aghaei2019learning,aghaei2020learning}. 
However, none of the existing MIP-based methods are able to efficiently handle constraints at \emph{path-level}, as the notion of ``path'' is not explicitly defined in the arc-based formulation.
Consider a use case where the sensitive feature is ``gender'' =\{M,F\}. For each rule $j$, we denote the gender-specific fairness metric as $\phi_{Mj}$ and $\phi_{Fj}$, and apply a path-level constraint: $|\phi_{Mj}-\phi_{Fj}|\leq \delta$, where $\delta$ refers to a user-specified bias tolerance. 
By applying the fairness constraint to each rule,  we eliminate blatantly unfair rules, which may be admissible in prior fairness-constrained MIP models that are unable to efficiently evaluate such path-level metrics. 

Meanwhile, a sample-level ``fairness budget'' constraint akin to \cite{aghaei2020learning} can also be modeled in our framework by summing the fairness metric over all paths,  i.e., $\sum_{j=1}^L|\phi_{Mj}-\phi_{Fj}|z_j\leq \delta$. 
Once path $j$ is delineated in the feature graph, the associated metrics can be defined and simply enter the RMP as coefficients associated with rule $j$.  For example, the cardinality constraint in the OMT formulation is a path-based restriction that directly controls the number of active leaf nodes in a tree, which is relatively difficult to enforce in prior MIP-based methods.

\section{Additional Details for Section 6: Experiments}
\subsection{Small/medium datasets}
\subsubsection*{Experiment setup}
We closely follow the experiment setup in \cite{aghaei2020learning}. 
As \textsf{FlowOCT} only allows binary input, we perform one-hot encoding on integer input and quantile discretization on features with real numbers. For \textsf{OMT}, we perform cumulative binning on numerical and ordinal features, where thresholds are determined via quantile discretization. Meanwhile, \textsf{OCT} and \textsf{BinOCT} use the original input. 

As \textsf{OCT} and \textsf{FlowOCT} have a regularization parameter  which controls the complexity of the tree, for each split and each depth, we tune the corresponding model by  selecting the regularization parameter from the set $\{0, 0.01, 0.1\}$ with the best performance on the validation set. Next, we retrain a model with this parameter on both the training and validation sets
and report the associated out-of-sample performance. 
The validation step is skipped for \textsf{BinOCT} which does not have such a tuning parameter and only constructs full trees with $2^d$ leaf nodes.  For \textsf{OMT}, we use cross-validation to  determine the number of discrete bins for numerical features. More specifically, the number of bins tested were \{3, 4, 5, 8\} and quantile discretization was applied to determine the corresponding thresholds.

\subsubsection*{Results}
Table \ref{table_small_data_full} shows the experimental results in terms of the out-of-sample accuracy averaged over five splits for $d\in\{2,3,4,5\}$,  conducted over the same 12 UCI datasets from that have been used in \textsf{FlowOCT} \cite{aghaei2020learning}, which is considered the state-of-the-art ODT. Best  accuracy in a given row is reported in \textbf{bold}.  Compared to three recently proposed MIP-based alternatives as well as CART, \textsf{OMT} dominates in terms of accuracy in 56.3\% of instances, \textsf{FlowOCT} 33.3\%, \textsf{OCT} 29.2\%, \textsf{BinOCT} 8.3\%, and \textsf{CART} 14.6\% (including ties). 

\begin{table}
 \centering
\begin{tabular}{rrrrr}
\toprule
 $d$ &    Mean &  Std. Dev &  Median  \\
\midrule
     2 & 0.08   &  0.05   &  0.10  \\
     3 &  0.59  & 2.21  & 0.14 \\
     4 & 4.37 & 10.35 & 0.21  \\
     5 & 3.65 & 6.94 & 0.26  \\
\bottomrule
\end{tabular} 
\caption{Optimality gap $\Delta$ $(\%)$}\label{optimality_gap}
\end{table}

\begin{table*}[t] 
\small
\caption{Mean ± standard deviation of out of sample accuracy on the small/medium datasets for $d\in\{2,3,4,5\}$. 
}
  \centering
\begin{tabular}{lllrccccc} 
\toprule
       dataset & $N$ & $d$ &         OMT &         OCT &      BinOCT &     FlowOCT &        CART \\
\midrule
 soybean-small &  47&    2 &  0.95±0.112 &   \B 1.0±0.048 & 0.972±0.048 &  \B 1.0±0.048 & 0.778±0.048 \\
 soybean-small &    47&  3 & 0.983±0.037 &   0.944±0.0 &    0.75±0.3 & 0.972±0.048 &    \B 1.0±0.0 \\
 soybean-small &    47&  4 & 0.883±0.139 & 0.944±0.048 &  0.833±0.22 & 0.944±0.083 &    \B 1.0±0.0 \\
 soybean-small &    47&  5 &  0.95±0.075 & 0.972±0.096 &  0.722±0.21 & 0.972±0.083 &   \B  1.0±0.0 \\
      monks-3 & 122&      2 & 0.967±0.013 & 0.966±0.004 & 0.966±0.004 & 0.966±0.004 & \B 0.971±0.012 \\
        monks-3 &    122&  3 & 0.987±0.008 & \B 0.99±0.011 &  0.99±0.011 & \B 0.99±0.011 & 0.986±0.019 \\
       monks-3 &    122&   4 & 0.987±0.008 &  0.99±0.015 & 0.988±0.011 &  0.99±0.011 & \B 0.993±0.007 \\
       monks-3 &    122&  5 & 0.987±0.008 & 0.978±0.014 & 0.983±0.015 &  0.99±0.012 & \B 0.993±0.007 \\
      monks-1 &   124&   2 & 0.751±0.035 & 0.751±0.041 & 0.751±0.041 & 0.751±0.041 & 0.734±0.019 \\
       monks-1 &    124& 3 &    \B 1.0±0.0 & 0.856±0.097 & 0.859±0.015 &  0.859±0.03 & 0.818±0.087 \\
       monks-1 &    124&  4 &    \B 1.0±0.0 &   \B 1.0±0.029 &     \B 1.0±0.0 &    \B 1.0±0.03 & 0.806±0.064 \\
       monks-1 &    124&  5 &     \B 1.0±0.0 & 0.935±0.142 &     \B 1.0±0.0 &   \B  1.0±0.03 & 0.787±0.042 \\
  hayes-roth &   132&   2 &  \B 0.615±0.11 & 0.608±0.014 &    0.55±0.1 & 0.608±0.052 &  0.425±0.09 \\
    hayes-roth &    132&   3 &  \B 0.78±0.111 & 0.758±0.038 &   0.7±0.115 &  0.725±0.09 & 0.517±0.038 \\
    hayes-roth &    132&  4 &  0.75±0.075 &  0.75±0.038 & 0.642±0.138 &   \B0.8±0.066 &  0.55±0.087 \\
    hayes-roth &132&       5 &  0.77±0.089 &  0.75±0.076 & 0.575±0.066 & \B 0.817±0.029 & 0.708±0.058 \\
        monks-2 &  169&    2 & 0.633±0.024 & \B 0.662±0.054 & 0.607±0.019 & \B 0.662±0.054 &  0.596±0.02 \\
       monks-2 &     169&   3 & 0.583±0.035 &  \B 0.662±0.04 & 0.585±0.044 & \B 0.662±0.043 & 0.594±0.034 \\
       monks-2 &     169&   4 & 0.603±0.045 & \B 0.662±0.031 & 0.581±0.027 & \B 0.662±0.023 & 0.598±0.033 \\
       monks-2 &     169&   5 & \B 0.779±0.03 &  0.662±0.05 &  0.607±0.03 & 0.662±0.055 & 0.651±0.055 \\   
  house-votes-84 &   232&   2 & \B 0.979±0.008 &  0.971±0.01 & 0.966±0.017 &  0.971±0.01 &  0.977±0.01 \\
house-votes-84 & 232&       3 & \B 0.976±0.009 &  0.971±0.01 &  0.971±0.01 &  0.971±0.02 &   0.96±0.02 \\
house-votes-84 &    232&   4 & 0.962±0.046 & \B  0.971±0.01 &  0.914±0.03 &  \B 0.971±0.01 &   0.96±0.02 \\
house-votes-84 &    232&   5 & 0.955±0.026 &  \B 0.971±0.01 &   0.96±0.01 &  \B 0.971±0.01 &   0.96±0.02 \\
         spect &   267&   2 & 0.755±0.054 & \B 0.836±0.071 & 0.781±0.071 & \B 0.836±0.071 & 0.711±0.023 \\
         spect &   267&   3 & 0.749±0.048 & \B  0.836±0.06 &  0.756±0.06 &\B 0.836±0.091 & 0.731±0.015 \\
         spect &  267&    4 & \B 0.834±0.04 & 0.801±0.086 & 0.746±0.065 & 0.791±0.079 & 0.731±0.026 \\
         spect &   267&  5 & \B 0.803±0.039 &  0.791±0.06 & 0.721±0.048 & 0.796±0.085 &  0.731±0.03 \\
  breast-cancer &    277&  2 &  \B 0.774±0.07 & 0.695±0.036 & 0.686±0.038 & 0.681±0.079 & 0.743±0.043 \\
 breast-cancer &    277&  3 &  \B  0.72±0.06 & 0.714±0.079 & 0.714±0.049 & 0.681±0.092 & 0.705±0.036 \\
 breast-cancer &    277&  4 &  0.74±0.046 & 0.724±0.038 & 0.662±0.091 & \B 0.743±0.079 &  0.69±0.058 \\
 breast-cancer &    277& 5 & 0.669±0.094 &   \B 0.738±0.0 & 0.567±0.157 & 0.676±0.036 & 0.671±0.049 \\
 balance-scale &  625&    2 & \B 0.703±0.043 & 0.665±0.019 &  0.65±0.013 & 0.671±0.007 &  0.62±0.016 \\
 balance-scale &    625&   3 & \B 0.726±0.006 & 0.696±0.051 & 0.665±0.042 & 0.696±0.026 & 0.709±0.037 \\
 balance-scale &    625&   4 & \B 0.781±0.016 & 0.747±0.048 & 0.707±0.006 &  0.699±0.01 & 0.769±0.033 \\
 balance-scale &    625&  5 & \B 0.772±0.012 &  0.735±0.01 &   0.565±0.1 &  0.72±0.045 & 0.762±0.013 \\
 tic-tac-toe &   958&   2 &  \B 0.71±0.018 &  0.689±0.05 & 0.689±0.021 &  0.689±0.05 & 0.672±0.035 \\
   tic-tac-toe &   958&   3 & 0.739±0.013 &\B 0.765±0.027 &  0.75±0.036 &  0.735±0.05 & 0.725±0.045 \\
   tic-tac-toe &   958&    4 & \B 0.802±0.039 & 0.776±0.073 & 0.786±0.021 & 0.757±0.032 & 0.758±0.019 \\
   tic-tac-toe &   958&    5 &  \B 0.82±0.061 & 0.711±0.025 & 0.812±0.029 &  0.788±0.05 & 0.778±0.046 \\
car-evaluation &    1728&  2 & 0.769±0.013 &  0.765±0.01 &  0.765±0.01 &  0.765±0.01 & \B 0.782±0.013 \\
car-evaluation &    1728&  3 & \B 0.803±0.015 &  0.789±0.02 & 0.786±0.029 & 0.798±0.016 & 0.782±0.033 \\
car-evaluation &    1728&  4 & \B 0.879±0.009 & 0.796±0.076 &  0.848±0.01 & 0.823±0.016 & 0.842±0.029 \\
car-evaluation &    1728&  5 &  \B 0.91±0.005 & 0.742±0.041 & 0.815±0.052 &   0.8±0.016 & 0.857±0.019 \\
      kr-vs-kp &   3196&   2 & \B 0.866±0.016 & \B 0.866±0.014 & \B 0.866±0.014 & \B 0.866±0.014 & 0.762±0.017 \\
      kr-vs-kp &    3196&  3 & \B 0.941±0.008 & 0.859±0.096 & 0.925±0.025 & 0.938±0.011 & 0.898±0.014 \\
      kr-vs-kp &    3196& 4 &  \B 0.96±0.006 & 0.847±0.094 & 0.938±0.012 &  0.94±0.011 &  0.94±0.011 \\
      kr-vs-kp &    3196&   5 & \B 0.968±0.003 & 0.652±0.098 & 0.847±0.164 & 0.946±0.057 &  0.94±0.011 \\
\bottomrule
\end{tabular}\label{table_small_data_full}
\end{table*}

Table~\ref{optimality_gap} reports on the MIP-LP gap with respect to $d$, which is defined as $\Delta = (\nu_{IP} - \nu_{LP})/\nu_{IP}$, i.e., the difference between the objective value achieved by the Master-MIP ($\nu_{IP}$) and the final RMP ($\nu_{LP}$). 
The small gaps suggest that the path-based LP relaxation is a relatively strong approximation of the nonconvex discrete OMT model. it also underscores the advantage of converging to a relatively small subset of high quality paths from which an effective decision tree can be distilled using a standard MIP solver.

{To illustrate the reduction in runtime and MIP size for large datasets, the CG procedure for the largest MIP-ODT instance analyzed in the literature in terms of sample size (\textit{Skin}) at $d = 2$, converges in $7\pm1$ minutes. The procedure generated less than $\hat{L} = 2400$ columns to improve upon the solution quality achieved by the multivariate MIP approach \textsf{S1O} that requires several hours and employs a data preprocessing step. This result is not surprising considering that this dataset yields a relatively small feature graph after discretizing its few numerical features. Hence, by Proposition 2, the number of feasible paths at $d = 2$ would be limited (and independent of data sample size). Prior MIP based binary ODT models are unable to exploit this hierarchical graph structure and employ an arc-based approach that requires more than $2^dN$ (~800,000) binary decision variables.} 

\subsection{Large datasets}
For the experiments with large-scale datasets, we down-sampled the original \textsf{susy} dataset to about a million samples to enable it to be processed on a laptop.

\clearpage
\bibliographystyle{plain}
\bibliography{ref}